\documentclass[sigconf]{acmart}
\usepackage{amsmath,amssymb,amsfonts}

\usepackage{graphicx}
\usepackage{textcomp}
\usepackage{xcolor}
\usepackage{booktabs}  
\usepackage{todonotes}
\usepackage{pgffor}
\usepackage{geometry}
\usepackage{caption}
\usepackage{subcaption}
\usepackage[acronym]{glossaries}
\glsdisablehyper
\makeglossaries 
\newacronym{reach}{REACH}{Recurrent Error-driven Adaptive Control Hierarchy}
\newacronym{dof}{DOF}{degrees of freedom}
\newacronym{drl}{DRL}{Deep Reinforcement Learning}
\newacronym{rl}{RL}{Reinforcement Learning}
\newacronym{cpg}{CPG}{Central Pattern Generator}
\newacronym{nef}{NEF}{Neural Engineering Framework}
\newacronym{osc}{OSC}{Operational Space Control}
\newacronym{spa}{SPA}{Semantic Pointer Architecture}
\newacronym{sd1}{strD1}{striatal D1}
\newacronym{sd2}{strD2}{striatal D2}
\newacronym{stn}{STN}{subthalamic nucleus}
\newacronym{gpi}{GPi}{globus pallidus internus}
\newacronym{gpe}{GPe}{globus pallidus externus}
\newacronym{lif}{LIF}{Leaky Integrate and Fire}
\newacronym[longplural={Rectified Linear Units}, shortplural={ReLUs}]{relu}{ReLU}{Rectified Linear Unit}

\newacronym[longplural={Exponential Linear Units}, shortplural={ELUs}]{elu}{ELU}{Exponential Linear Unit}
\newacronym[longplural={Multilayer Perceptrons}, shortplural={MLPs}]{mlp}{MLP}{Multilayer Perceptron}
\newacronym[longplural={Artificial Neural Networks}, shortplural={ANNs}]{ann}{ANN}{Artificial Neural Network}
\newacronym[longplural={Spiking Neural Networks}, shortplural={SNNs}]{snn}{SNN}{Spiking Neural Network}
\newacronym[longplural={Large Language Models}, shortplural={LLMs}]{llm}{LLM}{Large Language Model}
\newacronym{stdp}{STDP}{Spike-Timing-Dependent Plasticity}
\newacronym{nhil}{NHiL}{Neuromorphic-Hardware-in-the-Loop}
\newacronym{fts}{FTS}{Force-Torque Sensor}
\newacronym{ml}{ML}{Machine Learning}
\newacronym{hri}{HRI}{Human-Robot Interaction}
\newacronym{hpc}{HPC}{High-Performance Computing}
\newacronym{ai}{AI}{Artificial Intelligence}
\newacronym[longplural={Work Packages}, shortplural={WPs}]{wp}{WP}{Work Package}

\makeatletter
\newcommand{\Autoref}[1]{%
  \begingroup
  \renewcommand{\sectionautorefname}{Section}%
  \renewcommand{\figureautorefname}{Figure}%
  \renewcommand{\tableautorefname}{Table}%
  \autoref{#1}%
  \endgroup
}
\makeatother

\setcopyright{none}
\acmDOI{}

\acmConference[ICONS]{International Conference on Neuromorphic Systems}{2026}{Chicago, IL, USA}
\acmISBN{}

\begin{document}
\makeatletter
\renewcommand{\@makecaption}[2]{%
  \vskip\abovecaptionskip
  \sbox\@tempboxa{{\bfseries #1.} #2}%
  \ifdim \wd\@tempboxa >\hsize
    {\bfseries #1.} #2\par
  \else
    \global\@minipagefalse
    \hb@xt@\hsize{\hfil\box\@tempboxa\hfil}%
  \fi
  \vskip\belowcaptionskip}
\makeatother

\title[Spiking Humanoid Control]{A Spiking Neural Architecture for Coordinating \\ Arm and Locomotor Control}





\author{Lea Steffen}
\affiliation{%
  \department{Centre for Theoretical Neuroscience\\
  Dept. of Systems Design Engineering}
  \institution{University of Waterloo}
  \city{Waterloo}
  \country{Canada}}
\email{lsteffen@uwaterloo.ca}
\orcid{0000-0002-7485-6915}

\author{Kathryn Simone}
\affiliation{%
  \department{Centre for Theoretical Neuroscience\\
  Dept. of Systems Design Engineering}
  \institution{University of Waterloo}
  \city{Waterloo}
  \country{Canada}}
\email{kathryn.simone@uwaterloo.ca}
\orcid{0000-0002-3658-9377}

\author{Graeme Damberger}
\affiliation{%
  \department{Centre for Theoretical Neuroscience\\
  Dept. of Systems Design Engineering}
  \institution{University of Waterloo}
  \city{Waterloo}
  \country{Canada}}
\email{gdamberg@uwaterloo.ca}
\orcid{0009-0003-4280-2212}

\author{Travis DeWolf}
\affiliation{%
  \institution{Applied Brain Research}
  \city{Waterloo}
  \country{Canada}}
\email{travis.dewolf@appliedbrainresearch.com}
\orcid{0000-0002-8780-5631}

\author{Hudson Ly}
\affiliation{%
  \department{Dept. of Nanotechnology Engineering}
  \institution{University of Waterloo}
  \city{Waterloo}
  \country{Canada}}
\email{h6ly@uwaterloo.ca}
\orcid{0009-0001-6849-8082}

\author{Chris Eliasmith}
\affiliation{%
  \department{Centre for Theoretical Neuroscience \\ 
  Dept. of Systems Design Engineering \\
  Dept. of Philosophy}
  \institution{University of Waterloo}
  \city{Waterloo}
  \country{Canada}}
\email{celiasmith@uwaterloo.ca}
\orcid{0000-0003-2933-0209}

\renewcommand{\shortauthors}{Steffen et al.}

\begin{abstract}
\glspl{snn} coupled with neuromorphic hardware offer energy-efficient solutions for humanoid robot control. However, existing \gls{snn}-based motor control systems address bipedal locomotion and arm control in isolation, leaving integrated control of both unaddressed.
We present a spiking architecture that coordinates force-based arm control and bipedal locomotion in a simulated humanoid, using the \gls{nef} and \gls{spa}. High-level action selection between locomotor and arm control is mediated by a biologically grounded spiking basal ganglia model.
We validate the system through co-simulation of Nengo, for the neural control, and Isaac Sim, demonstrating successful target reaching, continuous digit drawing, path-following locomotion, and finally, switching between walking and arm control via basal ganglia disinhibition.
To our knowledge, this is the first integrated spiking controller to combine bipedal locomotion and arm control on a full-scale humanoid platform. The full spike-based implementation enables future deployment on low-power neuromorphic hardware.
\end{abstract}

\begin{teaserfigure}
  \centering
\includegraphics[width=\linewidth]{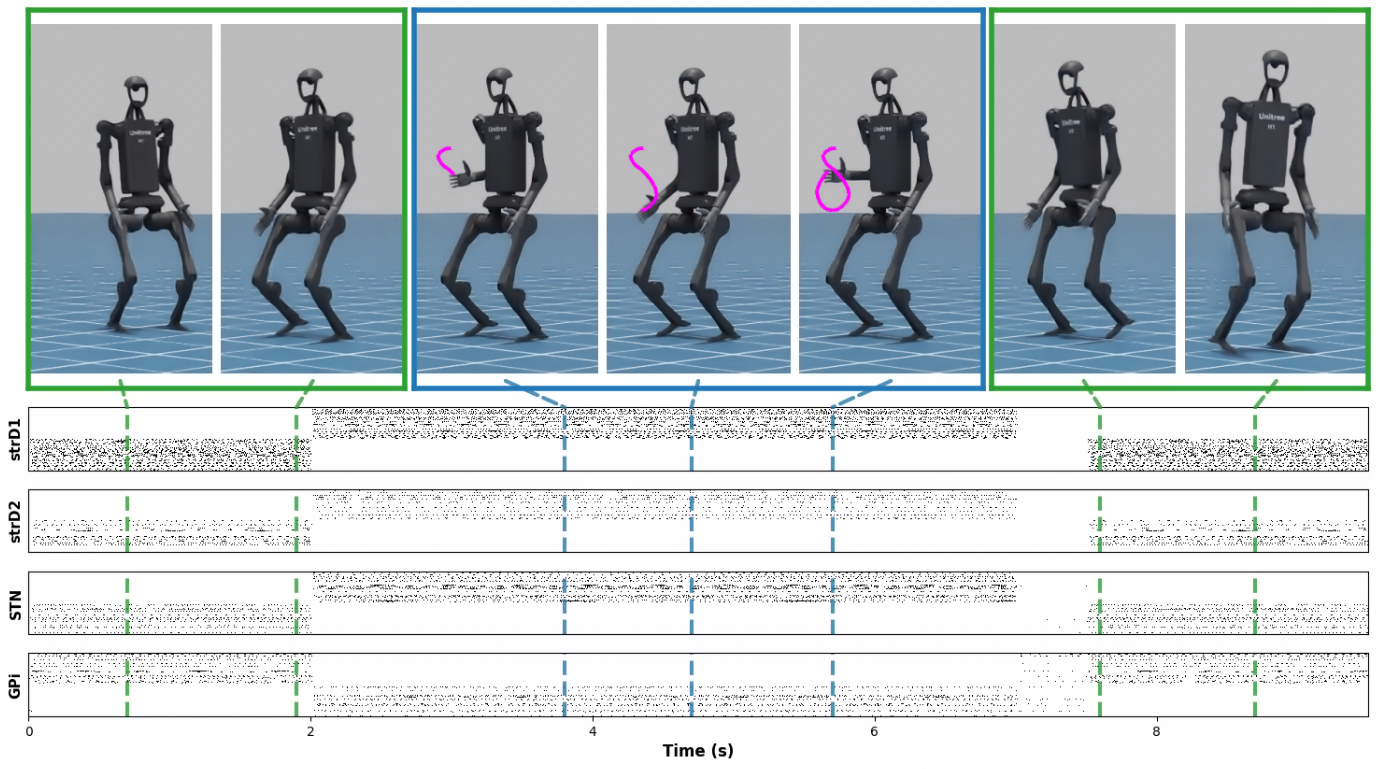}
\caption{The simulation snapshots show walking (green) and arm control for drawing the digit `8' (blue). Below are shown the spike trains of the basal ganglia (strD1, strD2, STN and GPi), which are used to coordinate these actions. Dashed lines link the robot states to their corresponding neural activity.}
  \label{fig:teaser}
\end{teaserfigure}

\maketitle

\section{Introduction} \label{sec:introduction}
\glsreset{snn} \glsreset{nef} \glsreset{spa}
\glspl{snn}, especially when coupled with neuromorphic hardware, promise low-power solutions to a wide variety of applications. Many of those applications are clustered around the abilities of biological systems, and demand fast, robust, and dynamic computation. Included in such applications is dynamic control of humanoid robots. Specifically, humanoid robots operating in human environments need a control system that can manage both the physics of walking and arm control. Yet, research in \gls{snn}-based motor control has largely evolved by treating such behaviours in isolation: existing systems achieve either bipedal locomotion or arm manipulation, but not both within a single integrated spiking architecture. 

Consequently, with an eye towards whole-humanoid control by a ~\gls{snn}, we present a unified spiking control architecture that integrates arm control and walking. Rather than treating these as separate concurrent tasks, our approach views them as an integrated capability essential for meaningful agent activities.
Our network employs a high-level spiking controller for action selection, which coordinates between locomotion and reaching by routing context-appropriate information to the lower-level modules. This allows the system to flexibly adapt its behavior based on the current task requirements.
 
Our contributions are as follows:
\begin{enumerate}
\item We demonstrate that the \gls{reach} model of biological arm control can be adapted to a full-scale humanoid platform, the Unitree H1 robot, demonstrating its utility on such a platform.
\item We propose a novel method for converting a pre-trained \gls{ann} locomotion policy to a spiking equivalent, in the case where neuron response functions in the \gls{ann} include negative activity.
\item We demonstrate, for the first time, integrated locomotion and manipulation within a single spiking network, achieved by applying the \gls{spa} methods to coordinate distinct locomotion and arm spiking controllers on the Unitree H1.
\end{enumerate}

The remainder of this paper is organized as follows. \Autoref{sec:related_work} surveys relevant prior work in neurorobotics, situating our contributions within the existing landscape of spiking motor control. \Autoref{sec:methodology} describes how all the modules interact to produce the behavior shown in \autoref{fig:teaser}; these modules are then explained in detail: the spiking arm controller based on the \gls{reach} model, the \gls{ann}-to-\gls{snn} conversion of the locomotion policy, and the \gls{spa}-based concurrent control architecture. \Autoref{sec:experiments} presents validation results, including arm control, locomotion, and action selection via the \gls{spa} switching mechanism, all evaluated on the Unitree H1 humanoid platform. \Autoref{sec:discussion} concludes with a discussion of the scientific contributions, their impact and limitations.

\section{Relevant Work in Neurorobotics} \label{sec:related_work}
Research reported in \cite{de_azambuja_diverse_2016} demonstrates spiking control on a humanoid-like platform using Liquid State Machines, where a BAXTER robot---a fixed-pedestal dual-arm cobot with no locomotion capability---learns to draw three 2D shapes using four joints. The approach relies on reservoir computing with a trained linear readout. Unlike the work presented here, this controller operates open-loop, involves no sensory feedback, and is not integrated with any other behaviour.

More broadly, individual subcomponents of humanoid motor control have been demonstrated with spiking networks in isolation: bipedal balancing \cite{petro_human_2022}, finger control of robotic hands via Hebbian learning on anthropomorphic actuators \cite{uleru_using_2022}, and head pose estimation through spiking path integration over the neck joints of the iCub robot \cite{polykretis_spiking_2022}, yet integrating locomotion with arm control, a prerequisite for multi-step physical interaction with the environment, remains unaddressed.

Vasquez Tieck et al. \cite{tieck_controlling_2018, tieck_multi-modal_2018, tieck_combining_2019} demonstrated \gls{snn}-based arm control using hierarchical motor primitives that avoid inverse kinematics through error-driven combinations of base and correction motions. Their compact networks (\textasciitilde 1,500 neurons per module) achieved target reaching through multi-modal activation patterns inspired by spinal cord circuitry. However, unlike the present work, these approaches focus on kinematic control without explicit force regulation, a key feature of biological arm control.

An alternative approach for solving robotic arm control is provided by the \gls{reach} model, which is based on research into the human motor control system. This method demonstrates excellent matches to biological arm control in monkeys and humans, and outperforms deep learning methods on reaching tasks. It was first presented in~\cite{dewolf_spiking_2016}, whereby a simplistic three-link arm is used for evaluation in simulation. The \gls{reach} model is also the cornerstone of the implementation of neuromorphic operational space control presented in \cite{dewolf_neuromorphic_2023}. That work describes position and orientation tracking of a 7 \gls{dof} robotic arm accomplished by running a \gls{snn} on Intel’s Loihi chip. We extend and apply this approach for the reaching component of our architecture.

Spiking control for legged robots is a rapidly emerging research area, providing an energy-efficient alternative to standard artificial neural network implementations. In \cite{rostro-gonzalez_cpg_2015} and \cite{lele_learning_2020}, a spiking implementation of a \gls{cpg} is used to control locomotion of a hexapod robot. 

In ~\cite{jiang_fully_2025} an \gls{snn} is combined with a with a policy network to control locomotion of a two-legged robot in an Isaac Gym simulation on various terrains.
The authors state that their system, implements spiking \gls{drl} and drastically improved robotic efficiency and control speed over traditional \glspl{drl}. The system achieved a 96\% reduction in power consumption, lowering the A1 robot's energy use from $500 J$ to just $20 J$ over a two-hour continuous test. Crucially, this efficiency translated to enhanced performance, controlling deviation ranges within 17\% across complex terrains and significantly increasing the robot's operational lifespan. However, a limitation is that the \gls{snn}-based \gls{drl} algorithm converges more slowly than the \gls{ann}-based equivalent, and for some robots, the initial training time was longer.
Locomotor control of a humanoid, specifically the Unitree H1 robot, sufficient to drive stable natural gait on flat terrain has been demonstrated with a \gls{mlp} deployed in a closed-loop control regime and trained in a \gls{rl} setup~\cite{isaac-sim_h1flatterrainpolicy_nodate}.
We adopt this approach and apply a novel ANN-to-SNN conversion method to demonstrate performance in a \gls{snn}.

Prior work in ANN-to-SNN conversion of \glspl{mlp}~\cite{eliasmith_large-scale_2012},~\cite{hunsberger_training_2016} suggests a possible path for direct approximation.
However, these previous approaches were applied on purely feedforward models, a network that is used in a closed-loop control regime. 
This demanded two additional enabling advances: a non-negative neural approximation of a signed activation function, to handle the network's hidden units that use the \gls{elu} non-linearity, and a spiking neuron representation of the movement command input sufficient to produce meaningful behavior.

\section{Methodology} \label{sec:methodology}
\begin{figure}[t!]
    \centering
\includegraphics[width=\linewidth]{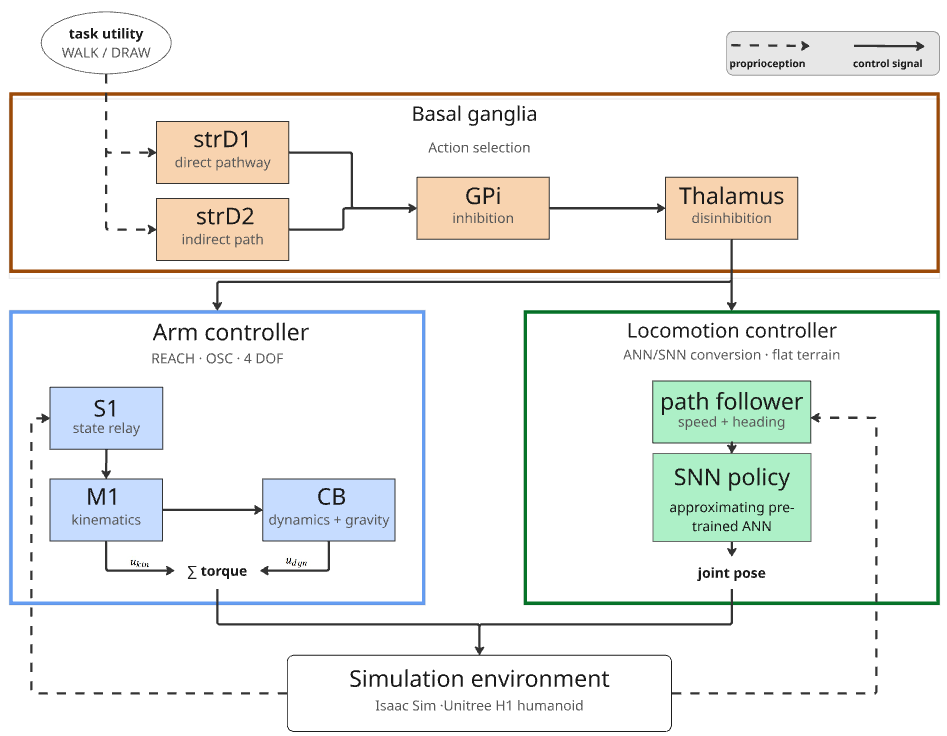}
    \caption{Overview of the integrated spiking motor control system. The basal ganglia (brown) performs action selection, gating either the arm controller (blue) or locomotion controller (green). Dashed lines indicate proprioceptive feedback; solid lines indicate control signals.}
    \label{fig:overview}
\end{figure}

The integrated architecture of our spiking motor control system for a humanoid robot is visualized in \autoref{fig:overview}. Task utility signals (`WALK' or `DRAW') are received by the basal ganglia module, where striatal populations (\gls{sd1}, \gls{sd2}) and the \gls{gpi} perform action selection via a direct and indirect pathway model. The \gls{gpi} output then projects to the thalamus, which disinhibits the selected action to gate either the arm controller or the locomotion controller.
An \gls{osc} based framework is used to control the right arm of the Unitree H1, which has 4 \gls{dof}. It includes three biologically grounded modules: the somatosensory cortex (\emph{S1}) which relays proprioceptive state to the primary motor cortex (\emph{M1}). \emph{M1} resolves the arm kinematics and baseline torques, sending an efference copy to the cerebellum (\emph{CB}), which computes corrective torques for dynamics and gravity compensation. These torques are summed and sent to the simulation environment.
The locomotion controller utilizes a \gls{snn} policy—approximating a pretrained \gls{ann}—to generate joint poses from speed and heading commands via a path follower. Both controllers output low-level motor commands (torques for the arm and joint poses for locomotion, respectively) to the simulation environment.

The arm control module is described in \autoref{sec:arm_control},
locomotion control in \autoref{sec:locomotion}, and finally
basal ganglia control mode mediation in \autoref{sec:combined_control}.

We are using the \gls{nef}~\cite{eliasmith_neural_2003} and the \glsreset{spa} \gls{spa}~\cite{eliasmith_how_2013} to implement motor control and decision making in spiking neurons.
The \gls{nef} acts as a neural compiler, determining connection weights to approximate target dynamics in spiking neurons, while the \gls{spa} proposes an architecture for building biologically constrained cognitive models, and includes methods for coordinating behaviour that we adopt here.
All modules of the spiking controller are implemented in Nengo~\cite{bekolay_nengo_2014}, a Python-based neural modeling platform that compiles models across hardware backends like CPUs, GPUs, FPGAs, or neuromorphic chips with minimal frontend changes. Its model library enables rapid construction of complex networks via reusable sub-networks. Specifically, we use NengoSPA, which is a Nengo-based implementation of the \gls{spa}.

\subsection{Spiking OSC for a 4 DOF Humanoid Arm} \label{sec:arm_control}
Our spiking arm control is based on the \gls{reach} model, mimicking the human motor control system~\cite{dewolf_spiking_2016, dewolf_neuromorphic_2023}.
\gls{reach} decomposes motor control across three cortical modules: 
primary somatosensory cortex \emph{S1}, primary motor cortex \emph{M1}, and cerebellum \emph{CB}. \emph{S1} acts as a sensory relay, encoding proprioceptive feedback for downstream use. \emph{M1} computes the kinematic component of an \gls{osc} signal, mapping task-space error to joint torques via the arm Jacobian and inertia matrix. \emph{CB} handles dynamics compensation, computing velocity-dependent damping torques and, in this extension, gravity compensation.
Together, these components are used to generate the force control signal sent to the arm, as formalized in \autoref{eq:u_gen}.
\begin{equation}
    u(t)=u_{kin}+u_{dyn}+u_g
    \label{eq:u_gen}
\end{equation}
All three modules are implemented in spiking neurons, specifically the \gls{lif} neuron model. \autoref{tab:modules_reach} summarises each module's biological analogue, inputs, and outputs.
\begin{table}[h!]
\centering
\renewcommand{\arraystretch}{1.6}
\begin{tabular}{p{.3cm} p{2.cm} p{2.cm} p{1.1cm} p{1.5cm}}
\toprule
\textbf{} & \textbf{Analogue} & \textbf{Function} & \textbf{Input} & \textbf{Output} \\
\midrule
\textbf{S1} & Primary \newline Somatosensory \newline Cortex & State \newline estimation &
  $[q,\, \dot{q},\, x_{ee}]$ & $[q,\, \dot{q},\, x_{ee}]$ \\[6pt]
\textbf{M1} & Primary \newline Motor Cortex & Kinematic + \newline inertia \newline transform &
  $[q,\, \tilde{x}]$ & $u_{kin}$ \\[6pt]
\textbf{CB} & Cerebellum & Dynamics \newline compensation \newline + gravity &
  $[q,\, \dot{q}]$ & $u_{dyn}$ \\
\bottomrule
\end{tabular}
\vspace{4pt}
\caption{Module overview with biological analogues, functions in the controller, as well as input and output.}
\label{tab:modules_reach}
\end{table}
%
Specifically, the module \emph{S1} encodes the current joint positions $q\in \mathbb{R}^{n_{DOF}}$, joint velocities $\dot{q}\in \mathbb{R}^{n_{DOF}}$, and end-effector position $x \in \mathbb{R}^{OS_{dim}}$, giving a total feedback vector of dimension $2 \cdot n_{DOF} + OS_{dim}$. The module is implemented using one independent ensemble per dimension, since \emph{S1} performs no nonlinear computation and each dimension can therefore be represented independently. The changes from the original \gls{reach} framework are minimal and accommodate 3D end effector feedback as shown in \autoref{tab:modules_reach}.

The module \emph{M1} receives the desired end-effector displacement, i.e., the difference between the target and the current end-effector position, and converts it into joint torques via the kinematic component of the \gls{osc} signal:   
\begin{equation} u_{kin} = J(q)^T \cdot M_x(q) \cdot k_p \cdot \tilde{x} 
\end{equation} 
where $k_p >0$ is the proportional controller gain vector for the task-space error $\tilde{x}$ (defined as $\tilde{x} = x_{\text{des}} - x_{ee}$).
This computation requires the arm's Jacobian $J(q)$ and inertia matrix $M_x(q)$, both of which depend non-linearly on the current joint configuration $q$. 
While the original \gls{reach} framework used a planar two-link arm~\cite{dewolf_spiking_2016}, extending the model to a 4-DOF arm operating in 3D space significantly increases the neural representational load. 
In a monolithic implementation, a single ensemble of neurons would need to encode both $q$ and the task-space error $\tilde{x}$ jointly. For a 4-\gls{dof} system in 3D, this results in a 7D input space ($q \in \mathbb{R}^4$ and $\tilde{x} \in \mathbb{R}^3$).
Such high dimensionality is notoriously difficult for spiking neurons to handle without large amounts of data to optimize the system.
%
We therefore exploit the linearity of the control equation in $x_{des}$, decomposing it exactly into three independent terms, one per axis $x, y, z$. By decomposing the computation into three independent spatial components, 
we can use three separate 5D ensembles ($q \in \mathbb{R}^4$ plus one scalar dimension of $\tilde{x}$) rather than one 7D ensemble: 
\begin{equation} 
u_{kin} = k_p \sum_{d=1}^{3} \tilde{x}_d \cdot f_d(q) \qquad \text{where } f_d(q) = J(q)^T M_x(q) 
\end{equation}
This decomposition enables the optimizer to more efficiently determine the required neural connection weights to compute this function.

The module \emph{CB} receives the current joint positions $q$ and velocities $\dot{q}$, and computes the dynamic component of the \gls{osc} signal. In particular, the velocity-dependent damping torques required to compensate for the arm's inertia:
\begin{equation}
u_{dyn} = M(q) \cdot k_v \cdot \dot{q} 
\end{equation}
where $M(q) \in \mathbb{R}^{4 \times 4}$ is the joint-space mass matrix and $k_v$ is a velocity damping gain. In contrast to~\cite{dewolf_spiking_2016}, which operated on a planar robot, the \emph{CB} also outputs a gravity compensation term:
\begin{equation}
u_g = -g(q) 
\end{equation}
Both $M(q)$ and $g(q)$ depend non-linearly on the joint configuration $q$, requiring the ensemble to jointly represent $[q, \dot{q}] \in \mathbb{R}^{8}$. This induces an 8D input space for a 4-\gls{dof} arm. As with \emph{M1}, this dimensionality is challenging for finding weights, and represents a significant increase over the two-link planar arm used in the original \gls{reach} framework, where the equivalent ensemble was only 4D.
Analogously to \emph{M1}, we exploit the fact that the dynamics equation is linear in $\dot{q}$, allowing an exact decomposition into a sum of $n_{DOF}$ independent terms, one per joint velocity:
\begin{equation}
u_{dyn} =  -k_v \sum_{j=1}^{4} \dot{q}_j \cdot M(q)
\end{equation}
This allows four separate 5D ensembles ($q \in \mathbb{R}^4$ plus one scalar $\dot{q}$) to replace the single 8D ensemble, each decoding $ \dot{q}_j \cdot M(q)$, with the final torque accumulated by summation. Gravity compensation $-g(q)$ depends only on $q$ and is handled by a dedicated 4D ensemble without further decomposition.


\subsection{Spiking Bipedal Locomotion} \label{sec:locomotion}
Our bipedal locomotion controller walks along a pre-defined path through the environment via a hierarchical control architecture:
The reinforcement-learning policy network optimized for locomotion on flat terrain, computes and sets the joint angles that continuously shape the robot's pose to produce walking. 
The policy takes as input both the robot's state and command signals specifying the desired forward speed and angular turn rate. These command signals are generated by a spiking path-following controller, which computes the error between the reference point on the path at time, and the robot's actual position.
We next describe these modules in turn, beginning with the low-level control policy, and then present the path-following controller.

\subsubsection{Spiking Control Policy for Flat Terrain Walking}
Our spiking locomotion control network is based on a pre-trained flat-terrain policy artificial neural network available through IsaacSim and associated PyTorch code~\cite{isaac-sim_h1flatterrainpolicy_nodate}. 
It is a 3-layer \gls{mlp} that provides joint positions for walking the robot Unitree H1 by mapping from a 69-dimensional input space to a 19-dimensional action space that defines the robot pose. 
The input is a concatenation of a high-level movement command, proprioceptive and kinematic state variables, and the network's last action. 
Internally, the architecture has three fully connected hidden layers, each with 128 nonlinear units.
Each hidden unit applies the \gls{elu} nonlinearity~\cite{clevert_fast_2015} and the network's last action is made available at the input through a recurrent connection from the output.

For hidden unit $i$ of layer $L$, the pre-activation is $z_{L,i}=\left\langle \mathbf{W}_{L,i},\mathbf{h}_{L-1}\right\rangle+b_{L,i}$, where $\mathbf{W}_{L,i}\in\mathbb{R}^{N_{L-1}}$ is the vector of weights from the previous layer, $b_{L,i}\in\mathbb{R}$ is the scalar bias, and $\mathbf{h}_{L-1}$ denotes the activity vector of layer $L-1$. The output of the unit is
\begin{equation}
h_{L,i} = \mathrm{ELU}(z_{L,i}) =
\begin{cases}
z_{L,i}, & z_{L,i} > 0, \\[4pt]
\alpha \bigl(e^{z_{L,i}} - 1\bigr), & z_{L,i} \le 0,
\end{cases}
\end{equation}
with $\alpha=1$ in the implementation.

As the \gls{elu} nonlinearity produces negative activations by design, this is not biologically feasible as the activity of real neurons is strictly positive. 
We therefore took this pre-trained policy network as a reference model and approximated each \gls{elu} hidden unit with a small ensemble of spiking neurons. Because this replacement requires spike-based communication between layers, synaptic filtering is introduced on feedforward connections, which in turn introduces an explicit dependence on time through layerwise propagation delays that are not in the original network. 
We therefore first describe the neural architecture used to approximate a scalar \gls{elu} hidden unit, and then describe how these approximations are assembled into a spiking multilayer locomotion controller.

\paragraph{Ensemble Approximation of the \gls{elu} nonlinearity.}
We require a non-negative approximation of $h_{L,i}$, denoted $\hat h_{L,i}$. The Neural Engineering Framework provides tools for finding a solution of the form $\hat h_{L,i}=\sum_{k=1}^N a_k d_k$, where $a_k$ is the activity of the $k^{\textsuperscript{th}}$ neuron of ensemble $i$ and $d_k$ is a scalar decoder weight obtained by optimization~\cite{eliasmith_neural_2003}. Motivated by the piecewise structure of the \gls{elu}, we explored a hybrid of \gls{lif} and \gls{relu} activation functions to approximate the function in its negative and positive halfspaces, respectively. 
This is natural because the \gls{elu} is continuous but qualitatively distinct on either side of the $z=0$ boundary: it is exactly \gls{relu}-like for $z>0$, while in the negative halfspace it is not functionally identical to an \gls{lif} response but shares its saturating asymptotic form.

Because the \gls{elu} is piecewise-defined, we approximated it by the sum of two ensembles, one assigned to the negative halfspace of pre-activations and the other to the positive halfspace:
\begin{equation}
\hat h_{L,i}(z_{L,i})
=
\sum_{n=1}^{N_n} a_n(z_{L,i}) d_n
+
\sum_{p=1}^{N_p} a_p(z_{L,i}) d_p.
\label{eq:elu_additive_approx}
\end{equation}
For this additive construction to remain accurate near the \gls{elu} knee at $z_{L,i}=0$, it is important that the two ensembles be gated by the sign of the pre-activation and that the summed output satisfy the boundary condition $\hat h_{L,i}(0)=h_{L,i}(0)=0$. More precisely, we desire that the negative-halfspace ensemble be inactive for non-negative pre-activations and that the positive-halfspace ensemble be inactive for non-positive pre-activations, that is, $a_n(z)=0$ for $z\ge 0$ and $a_p(z)=0$ for $z\le 0$.

Within the Neural Engineering Framework, the activity of a neuron in response to scalar stimulus $z$ is given by $a(z)=G\!\left[\alpha e z+\beta\right]$, where $G$ is the neuronal nonlinearity, $e\in\{-1,+1\}$ is the encoder, $\alpha$ is the gain, and $\beta$ is the bias. Since $z_{L,i}$ is one-dimensional, halfspace selectivity is imposed by fixing $e_n=-1$ and $e_p=+1$. The boundary condition at $z=0$ is then enforced by choosing the biases so that both branches are exactly at threshold at the origin, yielding $a_n(0)=a_p(0)=0$. 
In our implementation, this is achieved for the negative-halfspace LIF ensemble by selecting biases such that the input current at $z=0$ equals the threshold current, and for the positive-halfspace spiking rectified linear ensemble by setting the bias to zero. These choices are critical, since without them the sum of the two branch approximations would generally introduce a non-zero offset or overlapping contribution near $z=0$, degrading the approximation precisely at the transition point of the \gls{elu}.

The decoder weights associated with the two terms in Eq.~\eqref{eq:elu_additive_approx} are obtained by solving separate least-squares optimization problems on the negative and positive halfspaces:
\begin{align}
\tilde{\mathbf{d}}_n
&=
\arg\min_{\mathbf{d}_n}
\sum_{m=1}^{M_n}
\left(
\sum_{n=1}^{N_n} a_n\!\left(z^{(m)}_{L,i}\right)d_n
-
\mathrm{ELU}\!\left(z^{(m)}_{L,i}\right)
\right)^2,
\qquad
z^{(m)}_{L,i} \le 0, \\
\tilde{\mathbf{d}}_p
&=
\arg\min_{\mathbf{d}_p}
\sum_{m=1}^{M_p}
\left(
\sum_{p=1}^{N_p} a_p\!\left(z^{(m)}_{L,i}\right)d_p
-
\mathrm{ELU}\!\left(z^{(m)}_{L,i}\right)
\right)^2,
\qquad
z^{(m)}_{L,i} > 0.
\end{align}
The resulting approximation over the full domain is then given by Eq.~\eqref{eq:elu_additive_approx} with decoder vectors $\mathbf{d}_n=\tilde{\mathbf{d}}_n$ and $\mathbf{d}_p=\tilde{\mathbf{d}}_p$.

\paragraph{Reference Policy Approximation in a Spiking Network.}
Each \gls{elu} hidden unit of the reference policy is replaced by the sum of a negative-halfspace \gls{lif} ensemble and a positive-halfspace spiking rectified linear ensemble. In the spiking policy network, the scalar pre-activation for unit $i$ in layer $L$ is formed by a weighted connection from the previous layer together with an added scalar bias,
\begin{equation}
z_{L,i}(t)
=
\left\langle \mathbf{W}_{L,i}, \mathbf{h}_{L-1}(t) \right\rangle + b_{L,i},
\end{equation}
and this scalar drives the corresponding \gls{elu}-approximating subnetwork. 
Assembling these approximations across all hidden units yields a spiking surrogate of the original flat-terrain policy with the same layer widths, but with each scalar \gls{elu} unit replaced by a distributed spiking representation specialized to the negative and positive halfspaces of its pre-activation.

A challenge in this approximation strategy is that it introduces explicit temporal dynamics into what was a static layerwise computation. 
In other words, it is not sufficient to approximate the static \gls{elu} nonlinearity of each hidden unit in isolation. 
Rather, the full spiking policy must preserve the computation of the reference controller under the additional dynamical constraints imposed by interlayer spike-based communication.
In the reference policy, the network as a whole computes an algebraic transformation of its input within a control step, so that the network output at a given physics timestep is determined directly by the observation presented at that timestep together with the recurrently provided previous action. 
In the spiking policy, by contrast, communication between layers is mediated by spikes, so that each layer evolves as part of a discrete-time approximation to a continuous-time dynamical system.
To stabilize spike-based communication between layers, feedforward connections are filtered with a synapse of time constant $\tau_{\mathrm{fwd}} = 2\,\mathrm{ms}$. 
We found that this filtering provides sufficient temporal smoothing for reliable transmission of population activity while avoiding delays large enough to substantially degrade fidelity to the reference policy. 

\subsubsection{Path Following Controller}
We demonstrate spiking-walking control using a path-following task.
Consequently, we also provide a spiking implementation of a path-following controller that computes the error between the desired point along the path at timestep $t$, $s_{\mathrm{ref}}(t)$, and the actual robot state to generate a desired forward speed and angular rotation. In our implementation, the reference input is a 4-vector containing desired planar position and planar velocity, while the robot state is a 5-vector containing heading, angular velocity, planar position, and planar velocity:

\begin{equation}
\mathbf{s}_{\mathrm{ref}}(t) =
\begin{bmatrix}
x_{\mathrm{ref}}(t) \\
y_{\mathrm{ref}}(t) \\
\dot{x}_{\mathrm{ref}}(t) \\
\dot{y}_{\mathrm{ref}}(t)
\end{bmatrix},
\qquad
\mathbf{s}(t) =
\begin{bmatrix}
\theta(t) \\
x(t) \\
y(t) \\
\dot{x}(t) \\
\dot{y}(t)
\end{bmatrix}.
\end{equation}

The planar tracking error represented by the controller is then 
\begin{align}
\tilde{\mathbf{s}}_{xy}(t) &=\mathbf{s}_{\mathrm{ref}}(t)-[x(t),\,y(t),\,\dot{x}(t),\,\dot{y}(t)]^\top \\
 &= [\tilde{x}(t),\,\tilde{y}(t),\,\tilde{\dot{x}}(t),\,\tilde{\dot{y}}(t)]^\top.
\end{align}
The desired heading is taken to be the direction of the planar position error, with 
\begin{align*}
\theta_{\mathrm{des}}(t)&=\mathrm{atan2}\left(\tilde{y}(t),\tilde{x}(t)\right) \textrm{and} \\
\tilde{\theta}(t)&=\mathrm{wrap}{[-\pi,\pi)}\left(\theta_{\mathrm{des}}(t)-\theta(t)\right) \textrm{, where} \\
\mathrm{wrap}_{[-\pi,\pi)}(\phi)&=(\phi+\pi)\bmod 2\pi-\pi.
\end{align*}

The forward-speed command is computed from the magnitudes of the position and velocity errors, with 
\begin{align*}
\tilde{S}_p(t)&=\sqrt{\tilde{x}(t)^2+\tilde{y}(t)^2},  \\
\tilde{S}_v(t)&=\sqrt{\tilde{\dot{x}}(t)^2+\tilde{\dot{y}}(t)^2}, \textrm{ and}  \\
u_s(t)&=\max\left(0.01,k_{p,s}\tilde{S}_p(t)+k_{d,s}\tilde{S}_v(t)\right), \textrm{ with angular command} \\
u_\theta(t)&=k_{p,\theta}\tilde{\theta}(t)  \\
\end{align*}
In our implementation, the heading-error pathway provides proportional angular error only, such that the angular command is simply a proportional controller.
The controller output is therefore the 2-vector $\mathbf{u}(t)=[u_s(t),\,u_\theta(t)]^\top$, which is provided as input to the spiking policy network controller described previously.
\subsection{Integrated Control Architecture}  \label{sec:combined_control}

To coordinate the arm and locomotion controllers described in previous sections, we adopt the \gls{spa} approach to controlling interactions between networks. Specifically, we use the \gls{spa} spiking model of the basal ganglia~\cite{stewart_dynamic_nodate}, which is known to be critical for action selection in motor control.
The network suppresses all actions except the most valuable, which is disinhibited while all others are inhibited. Beyond its action selection function, the architecture mirrors the known organisation of the human basal ganglia, comprising five neurologically grounded ensembles.
These five ensembles, \gls{sd1}, \gls{sd2}, \gls{stn}, \gls{gpi} and \gls{gpe}, each instantiated per action, are interconnected following known neuroanatomical pathways. 
Input is distributed to \gls{sd1}, \gls{sd2} and \gls{stn} via scaled transforms. GABAergic projections ($\tau = 8 ms$) carry inhibition from striatum (\gls{sd2} and \gls{sd1}) to \gls{gpi} and \gls{gpe}, while the \gls{stn} drives broad excitation across both pallidal nuclei via AMPA synapses ($\tau = 2 ms$). \gls{gpe} feeds inhibition back to \gls{stn} and \gls{gpi}, and \gls{gpi}'s final inhibitory output, scaled by a negative output weight, enacts action selection.

In this framework, the \gls{sd1} ensembles serve as the primary input stage, where firing rates are driven by continuous value signals. The selection is realized through selective disinhibition: the \gls{gpi} serves as the final inhibitory output, maintaining a high baseline firing rate that suppresses all motor commands. When an action's value is high, the corresponding \gls{gpi} cluster is suppressed to near-zero spiking activity. This removes the inhibitory input from the downstream pathways, effectively releasing the selected motor behavior while the unselected actions remain inhibited.

\section{Validation on a Humanoid Platform} \label{sec:experiments}
All experiments are conducted in a co-simulation setup that couples Nengo and Isaac Sim. The latter serves as the physics engine, simulating the Unitree H1---a 19-\gls{dof} full-size humanoid platform---at 200 Hz, while our spiking neural controller runs in Nengo at 1000 Hz. The two simulators communicate via ROS 2,  exchanging state observations and control signals. A third orchestration layer, the \texttt{ExperimentManager}, manages this communication by synchronizing state and control signals across processes, scheduling command phases and handling data logging. Each experiment is defined by three files: an Isaac Sim script specifying the physical environment and robot configuration, a Nengo script implementing the spiking control architecture, and a test script that configures the command schedule, sets simulation parameters, and launches both simulators in coordination.

\subsection{Spiking Arm Control on the Unitree H1} \label{sec:experiments_arm}

\begin{figure}[h!]
    \centering
    \begin{subfigure}{\linewidth}
        \centering
        \includegraphics[width=\linewidth]{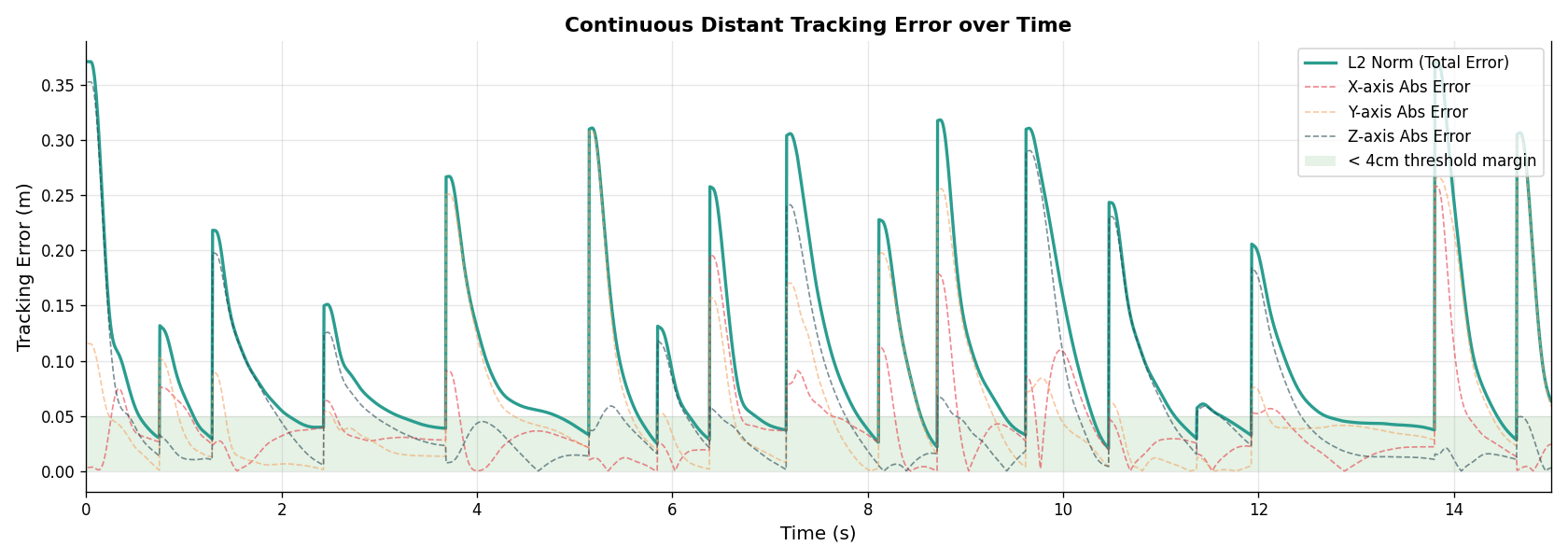}
        \caption{}
        \label{fig:tracking_error}
    \end{subfigure}
    
    \vspace{0.5em}
    
    \begin{subfigure}{\linewidth}
        \centering
        \includegraphics[width=\linewidth]{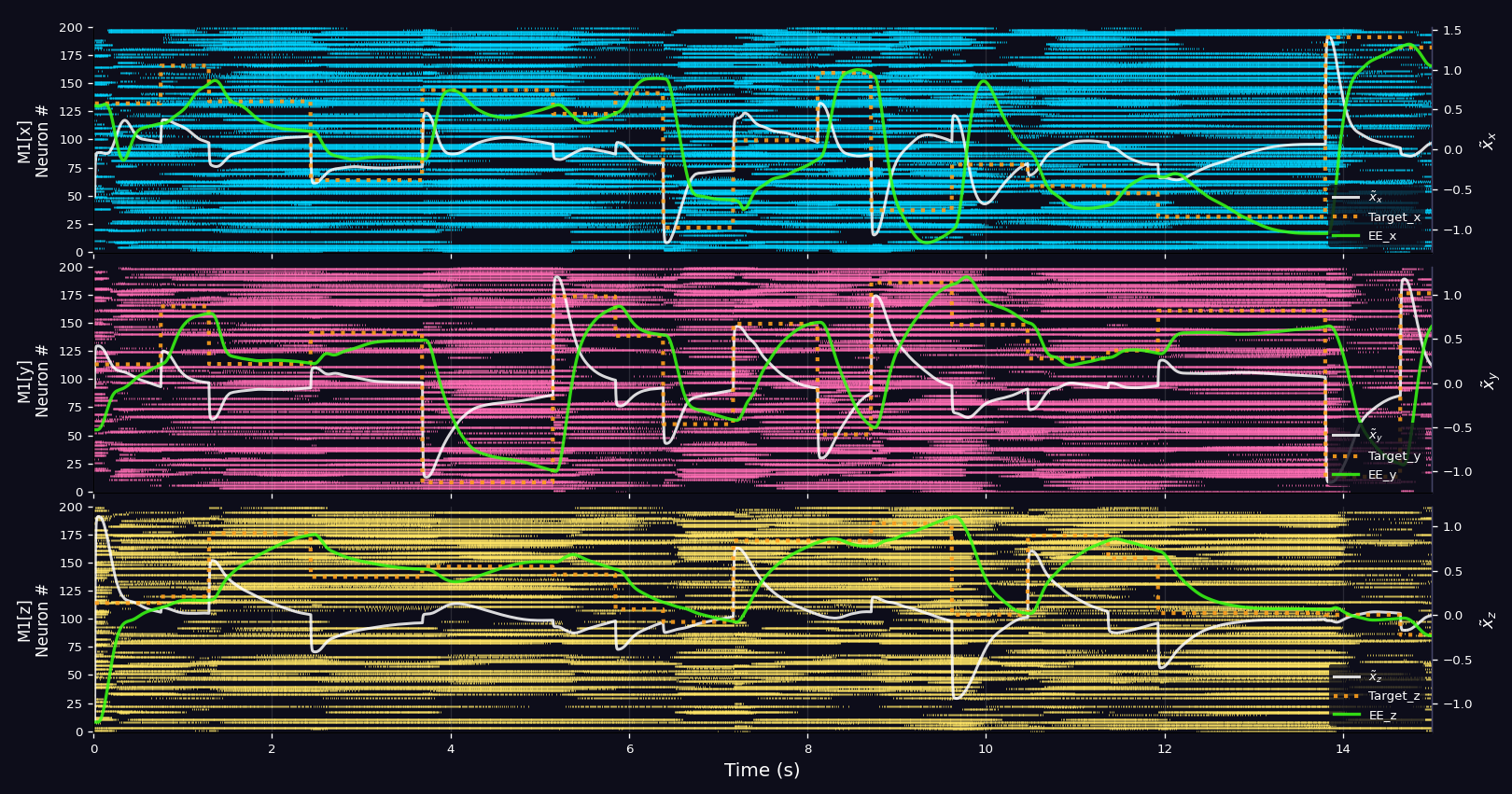}
        \caption{}
        \label{fig:Neural Dynamics_EE_Target}
    \end{subfigure}
    
    \caption{Reach-to-target task, if end-effector surpasses 4 cm threshold a new target is generated. (\subref{fig:tracking_error}) Continuous tracking error over time. The turquoise solid line represents the Euclidean distance between the robotic end-effector and the randomized spatial targets. 
    (\subref{fig:Neural Dynamics_EE_Target}) Neural spike raster plots and corresponding decoded variables for each operational axis ($X, Y, Z$). The physical end-effector trajectory (green), the error (white) and the continuously moving reference target (dotted orange) are overlaid.}
    \label{fig:target_follow}
\end{figure}
As a first task to evaluate the arm controller we use a simple reach-to-target task. A reach is considered successful if the position of the end-effector is within 4 cm of the target, after which a new target is generated randomly within defined area in front of the robot.

In \autoref{fig:tracking_error}, the continuous Euclidean tracking error is visualized over the duration of 15 s, during which 16 reaches are completed. The sharp peaks correspond with the instantaneous shifts in the target coordinate upon successful acquisition of the previous target, while the descending profiles demonstrate the spiking controller's ability to drive the arm to minimize the spatial discrepancy.
To highlight the system's responsiveness, in \autoref{fig:Neural Dynamics_EE_Target} the target is visualized as a dashed line in orange alongside the actual end-effector position (EE), shown in green. Overlaid on the left $y$-axis are the corresponding spike raster plots for the sub-populations of the \emph{M1} factored ensemble responsible for each spatial dimension, demonstrating dimension-specific neural activation. 

\begin{figure}[h!]
     \begin{subfigure}{0.5\linewidth}
         \centering
         \includegraphics[width=\linewidth]{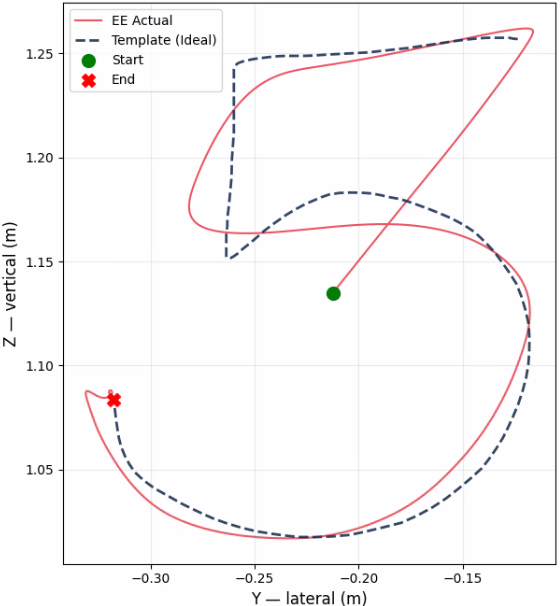}
         \caption{}
         \label{fig:nr5_2d_trajectory}
     \end{subfigure}
     \hfill 
     \begin{subfigure}{0.475\linewidth}
         \centering
         \includegraphics[width=\linewidth]{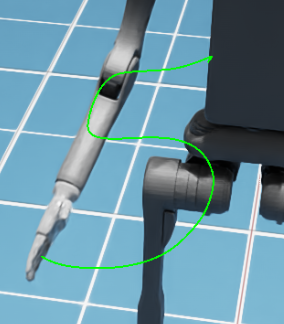}
         \caption{}
         \label{fig:nr5_isaacsim}
     \end{subfigure}
     \caption{Drawing the digit `5'. 
     (\subref{fig:nr5_2d_trajectory}) Resulting 2D end-effector trajectory (red) accurately tracing the reference template for the digit "5" (dashed).
     (\subref{fig:nr5_isaacsim}) Isaac Sim visualization of the Unitree H1 actively performing the drawing motion.}
     \label{fig:drawing5}
\end{figure}

To evaluate the continuous tracking capabilities of the adapted \gls{reach} controller, we implemented a digit drawing task. In this task, the right arm of the Unitree H1 is commanded to trace a predefined 2D trajectory representing a handwritten digit within the operational space. The reference trajectory is mapped onto the lateral ($Y$) and vertical ($Z$) axes relative to the robot's torso, with the depth ($X$) held constant. The target waypoints are continuously streamed to the spiking controller, which dynamically computes the required joint torques. This task tests the system's ability to maintain stable end-effector tracking under continuously changing velocity profiles and directional shifts.

In \autoref{fig:drawing5} the digit `5' is drawn. 
\Autoref{fig:nr5_2d_trajectory} shows the resulting 2D trajectory of the end-effector in the $Y,Z$ operational plane, demonstrating that the physical arm tracks the  reference template of the target digit (dashed). In \autoref{fig:nr5_isaacsim}, the Unitree H1 robot performs the commanded drawing motion within the Isaac Sim physics environment.

For all arm experiments (see \autoref{fig:teaser}, \ref{fig:target_follow}, \ref{fig:drawing5}), \emph{M1} uses
8,000 neurons per factored dimension,
\emph{CB} uses 200 neurons per factored dimension,
and \emph{S1} 50 neurons per ensemble. Therefore, 25,550 (24,000 + 1,000 + 550) spiking neurons actively drive the arm control pipeline.

\subsection{Spiking Locomotion on Flat Terrain} \label{sec:experiments_locomotion}

\begin{figure}[h!]
    \centering
    \begin{minipage}[b]{0.49\linewidth}
        \centering
        \begin{subfigure}{\textwidth}
            \centering
            \includegraphics[height=3.5cm, keepaspectratio]{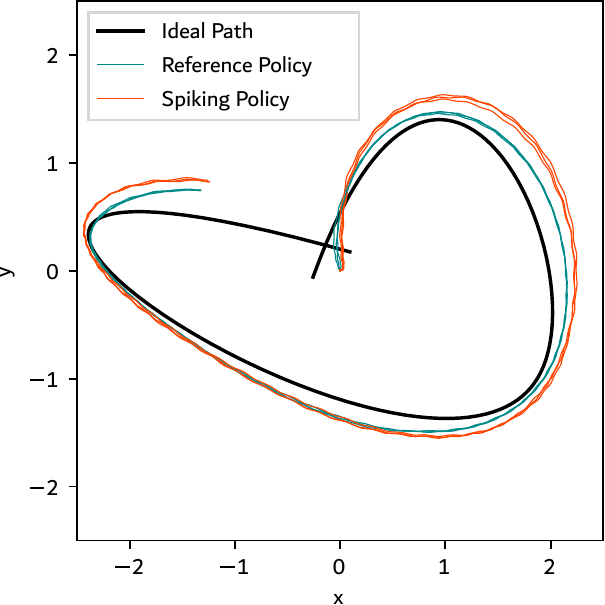}
            \caption{}
            \label{fig:sftp_panel_a}
        \end{subfigure}

        \vspace{0.5cm} 

        \begin{subfigure}{\textwidth}
            \centering
            \includegraphics[height=3.5cm, keepaspectratio]{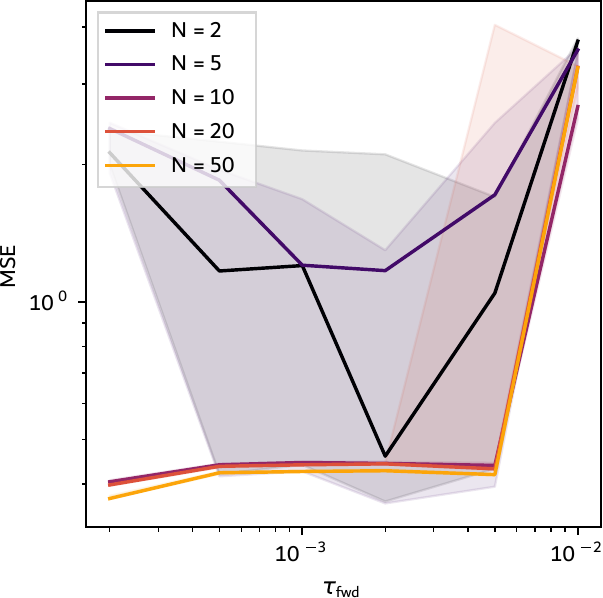}
            \caption{}
            \label{fig:sftp_panel_b}
        \end{subfigure}
    \end{minipage}
    \hfill
    \begin{minipage}[b]{0.49\linewidth}
        \centering
        \begin{subfigure}{\textwidth}
            \centering
            \includegraphics[height=8.cm, keepaspectratio]{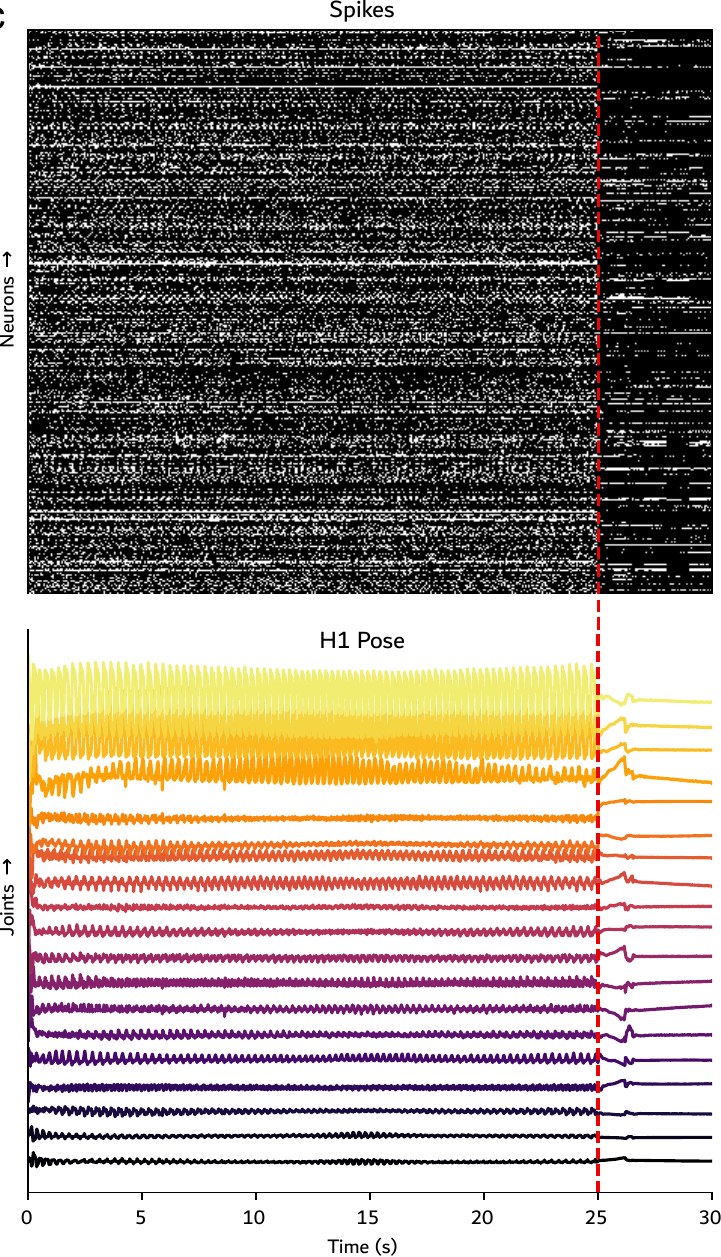}
            \caption{}
            \label{fig:sftp_panel_c}
        \end{subfigure}
    \end{minipage}

    \caption{Approximation of a trained reference policy for bipedal locomotion in spiking neurons
    (\subref{fig:sftp_panel_a}) Trajectories followed by Unitree H1 controlled by either the reference policy or its spiking approximation, across 5 randomly-selected seeds. Both networks received input from the spiking path following controller, conditioned on an ideal path through the terrain.
    (\subref{fig:sftp_panel_b}) Mean Squared Error (MSE) between the ideal and actual path of Unitree H1 with the spiking policy, as a function of the number of neurons per hidden unit in the reference policy ($N$), and time constants of synapses between layers ($\tau_{fwd}$). 
    (\subref{fig:sftp_panel_c}) The relationship between spiking activity and robot pose during bipedal locomotion. The red vertical line at $t = $ 25 seconds indicates the onset of inhibitory current injected into a randomly-selected 50\% of ensembles, resulting in a qualitative change in spiking activity and abrupt end to natural gait-associated oscillations.
    }
    \label{fig:locomotion}
\end{figure}
To validate our locomotion controller, we assessed its capacity to have the robot follow a simple pre-defined walking path through the environment. \Autoref{fig:locomotion}a depicts the trajectories followed by the Unitree H1 with respect to this ideal path (black) in two cases of interest: one in which the robot's pose was set by the reference policy \gls{ann} (teal), and one in which they were set by its \gls{snn} approximation (described in \autoref{sec:locomotion}; orange).
Performance was assessed through 5 experiments wherein the neural models were initialized with randomly-selected seeds, confirming the consistency of the results shown in \autoref{fig:sftp_panel_a}
This was necessary as in both cases, the spiking high-level path-following controller was always in the control loop.
The paths followed using the reference policy were all similar to the ideal path and to one another, demonstrating at once efficacy of the spiking path following controller and its integration with the reference policy, as well as robustness to the particular low-level spiking neuron parameters of the spiking path following controller.
However, the robot path approaches and cross at areas of high-curvature, consistent with a limited capacity to follow tight turns, and also ends prematurely near the point $x=-1,y=1$, indicative of a lag in the response to control error.
We then replaced the reference with its spiking approximation using $\tau_{fwd} = 2$ ms and used $N$ = 20 neurons in each \gls{elu}-approximating ensemble, split evenly between the negative and positive halfspaces.
This resulted in a set of paths highly similar to that produced with the reference, although small oscillations transverse to travel direction can be seen, revealing that while the spiking approximation was sufficient for stable walking behavior, natural gait was associated with larger perturbations of center of mass than those of the control policy.

\Autoref{fig:sftp_panel_b} characterizes performance of the spiking approximation as a function of its hyperparameters.
The plots suggest a hard upper limit of $\tau_{fwd}=5$ ms for reasonable performance, and a minimum of $N$ = 10 neurons for reliable good performance provided the earlier constraint is satisfied. 
Intriguingly, performance is strikingly consistent within this regime, demonstrating no benefit of additional neural resources beyond the $N$ = 10 level.
As a final verification, we assessed the causal role of spiking activity in locomotion control.
\Autoref{fig:sftp_panel_c} shows spiking activity (top) and the synchronously-collected oscillations of the robot pose associated natural gait (bottom), which surprisingly showed no obvious correlation in the side-by-side juxtaposition.
We therefore next assessed the necessity of the network by silencing a random 50\% of ensembles during the walking task.
The temporal evolution of robot pose abruptly changed, most notably in the cessation of oscillations observed during walking behavior.
This at once demonstrates a direct influence of the spiking network on robot pose, and their requirement maintaining stable natural gait. 
\subsection{Control via Basal Ganglia Action Selection} \label{sec:experiments_combined}
To validate the integrated architecture, the basal ganglia action selection is evaluated over a continuous, multi-phase simulation sequence consisting of four phases: an initial locomotion phase, the arm control phase, a brief arm-retraction transition phase, and a final locomotion phase.
\begin{figure}[h!]
    \centering
\includegraphics[width=\linewidth]{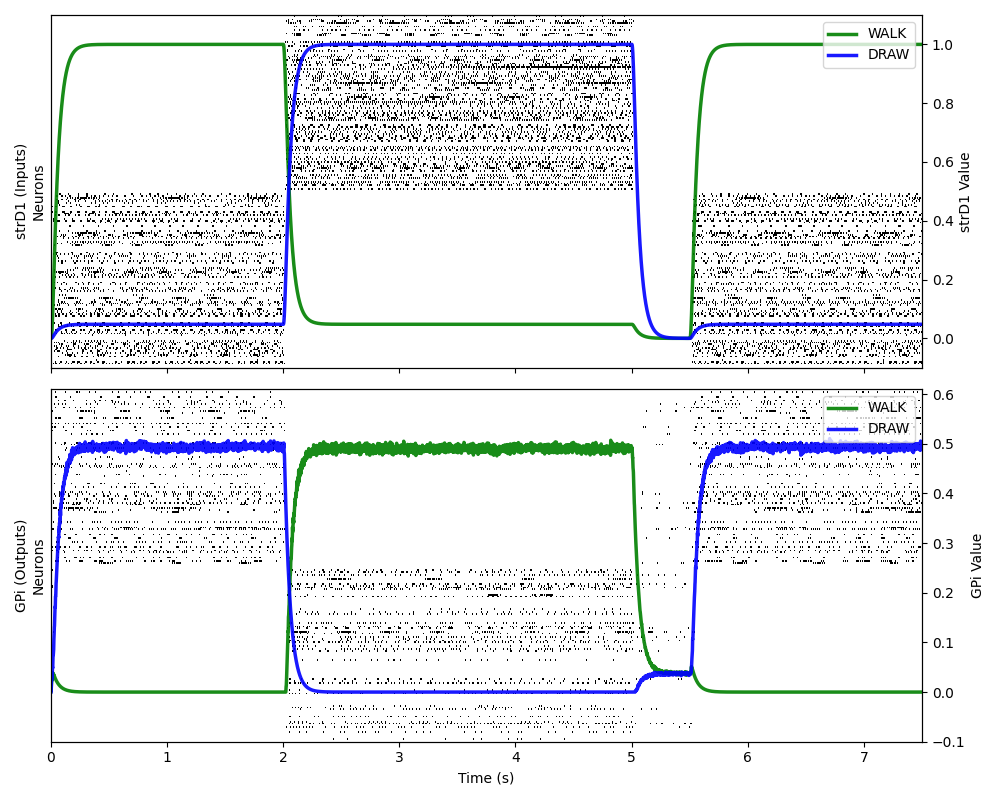}
    \caption{Spiking continuous action selection in the basal ganglia. (Top) Input command values overlaid with the spikes of striatal D1 (strD1) neurons, representing the excitatory drive for the `WALK' and `DRAW' actions over time. (Bottom) Action selection outputs overlaid on the globus pallidus internus (GPi) spiking raster.  Actions are selected through disinhibition, so high output activity indicates the action is not selected.}
    \label{fig:BG_input_output_spikes}
\end{figure}
 The transition phase is required to retract the arm to a safe configuration before resuming walking. The visualized arm control task is drawing the digit `8', but this can easily be swapped with a target-reaching task.
\autoref{fig:teaser} presents the full integrated sequence, showing simulation snapshots alongside synchronized basal ganglia spiking rasters (\gls{sd1}, \gls{sd2}, \gls{stn}, and \gls{gpi}). The dashed lines link physical robot states to their corresponding neural activity across all four phases. The basal ganglia successfully mediates this sequence. The arm-retraction transition is clearly identifiable as a change in spiking activity between the drawing and final locomotion task.

The temporal progression of the decision-making process at the neuronal level during a transition between walking and arm control is also visualized in \autoref{fig:BG_input_output_spikes}. 
The top panel displays the input stage, where the spiking activity of the \gls{sd1} ensembles closely tracks the overlaid utility trajectories for `WALK' and `DRAW'. As the `WALK' utility rises, we observe a temporally correlated surge in clustered excitation within the associated striatal neuronal population.
The bottom panel of \autoref{fig:BG_input_output_spikes} demonstrates the resulting selected action within the \gls{gpi}. Initially, the \gls{gpi} maintains a steady suppressive firing rate across both action dimensions. However, as the `WALK' utility dominates, the corresponding \gls{gpi} cluster exhibits a sharp cessation of spiking activity. This transition highlights the model's ability to autonomously resolve competition through disinhibition, cleanly releasing the `WALK' motor sequence while the `DRAW' program remains suppressed.

\section{Conclusion}  \label{sec:discussion}
In this work, we have demonstrated what we believe to be the first integrated controller capable of performing locomotion and arm control, and of switching smoothly between the two, in a spiking neural network. We combined and built upon previous work on these tasks to develop this controller. Implementing it fully in spikes enables it to be deployed at low power on neuromorphic hardware.

\subsection{Scientific Contribution and Impact}
We demonstrate the first integrated \gls{snn} to jointly coordinate bipedal locomotion and arm control on a full-scale humanoid, unifying walking, drawing, and state-switching within a single spiking architecture.

For arm control, the decomposition of high-dimensional kinematic and dynamic computations into independent lower-dimensional ensembles offers several advantages. Reducing the input space from 7D to three independent 5D ensembles significantly improves signal-to-noise ratio and decoding accuracy. This decomposition also aligns with the natural physics of the system, as torques generated by displacements in different spatial dimensions exhibit a degree of independence. It remains future work to determine if this decomposition is consistent with neural data.

For locomotion, the present work builds on the ANN-to-SNN conversion lineage established by Spaun~\cite{eliasmith_large-scale_2012} and subsequent vision work, extending it to a substantially different setting: rather than approximating a feedforward vision model, we approximate a deep control policy operating in closed-loop during continuous humanoid locomotion. This distinction is important because approximation errors are not confined to static input-output mappings but are amplified over time through body-world interactions. Unlike prior work, we re-use trained reference network parameters and extend this to a signed hidden-unit nonlinearity — to our knowledge, the first conversion of an ELU-based control policy. Our method is best understood as a three-step conversion strategy: training a non-spiking policy via conventional reinforcement learning, approximating it in a spiking substrate through convex optimization of neural decoders, and tuning the synaptic filter for control accuracy. This distinguishes our approach from direct training methods such as PopSAN~\cite{tang_deep_2020}, Proxy Target~\cite{xu_proxy_2025}, and SpikeRL~\cite{tahmid_spikerl_2026}. The successful non-negative approximation of the \gls{elu} nonlinearity suggests the strategy may generalise to other signed nonlinearities, and its systematic exploration across continuous control settings is a promising direction for future work.

\subsection{Limitations}
While arm control, locomotion, and action selection are fully spike-based, the system’s postural stability during arm control is not yet emergent from neural dynamics. Instead, stability is enforced by grounding the pelvis to the world via an architectural anchor, which serves as a placeholder for a future spiking balance-control module. Integrating bio-inspired spiking models to achieve dynamic balance control remains for future work.

Regarding locomotion, several open questions remain. We did not compare the adopted parameter-preserving conversion strategy against the layer wise approximation approach used in Spaun's visual system. It also remains unknown whether the hybrid \gls{relu}/\gls{lif} ensembles are strictly necessary, or whether a homogeneous ensemble of \gls{lif} neurons alone would have been sufficient. Finally, the present hybrid ensemble may be less suitable for signed nonlinearities whose negative halfspace behavior is more structured than the \gls{elu}'s simple saturation, though this remains to be tested.
The current system is further limited to completely flat terrain, as the converted locomotion policy was trained on this basis. However, the same conversion approach could be extended by adapting a more complex pre-trained walking policy, enabling robust spiking locomotion control across non-flat terrains.

\bibliographystyle{ACM-Reference-Format}
\bibliography{references}

\end{document}